\documentclass[11pt,a4paper]{article}
\usepackage[hyperref]{acl2019}
\usepackage{graphicx}  
\usepackage{times}
\usepackage{latexsym}
\usepackage{multirow}
\usepackage{textcomp}

\usepackage{url}
\usepackage{breakurl}
\usepackage[utf8]{inputenc}
\usepackage{graphicx}
\usepackage{array}
\usepackage{amsmath}
\usepackage[font=singlespacing]{caption}
\usepackage{fancyhdr}
\usepackage{amssymb}
\usepackage{nccmath}
\usepackage{color}
 \usepackage[normalem]{ulem}
 \usepackage{lipsum}
 \usepackage{graphics}
 \usepackage{tikz-dependency}
 \usepackage{enumitem}
\usepackage{tikz-qtree}
\usepackage{tikz}
 \usepackage{pgfplots}
  \usepackage{pgfplotstable}
  \usepackage{relsize}
  \usepackage[normalem]{ulem}

\usepackage{url}

\aclfinalcopy 
\def\aclpaperid{2017} 

\newcommand{\ignore}[1]{}
\newcommand{\E}{\mathbb{E}}

\newenvironment{itemizesquish}{\begin{list}{\labelitemi}{\setlength{\itemsep}{0em}\setlength{\labelwidth}{0.5em}\setlength{\leftmargin}{\labelwidth}\addtolength{\leftmargin}{\labelsep}}}{\end{list}}

\DeclareMathOperator*{\softmax}{softmax}

\DeclareMathOperator*{\argmin}{arg\,min}

\title{Scalable Syntax-Aware Language Models Using Knowledge Distillation}

\author{Adhiguna Kuncoro$^{\spadesuit\diamondsuit}$ ~ Chris Dyer$^{\spadesuit}$  ~ Laura Rimell$^{\spadesuit}$\\\textbf{Stephen Clark}$^{\spadesuit}$ \textbf{Phil Blunsom}$^{\spadesuit\diamondsuit}$ \\
$^{\spadesuit}$DeepMind, London, UK \\
$^{\diamondsuit}$Department of Computer Science, University of Oxford, UK\\
{\small \tt \{akuncoro,cdyer,laurarimell,clarkstephen,pblunsom\}@google.com}
}

\date{}

\begin{document}
\maketitle
\begin{abstract}
Prior work has shown that, on small amounts of training data, 
syntactic neural language models 
learn structurally sensitive generalisations 
more successfully than 
sequential language models. However, their computational complexity renders scaling difficult, and it remains an open question whether structural biases are still necessary when sequential models have access to ever larger amounts of training data. 
To answer this question, we introduce an efficient knowledge distillation (KD) technique that transfers knowledge from a syntactic language model trained on a small corpus to an LSTM language model, hence enabling the LSTM to develop a more structurally sensitive representation of the larger training data it learns from. On targeted syntactic evaluations, we find that, while sequential LSTMs perform much better than previously reported, our proposed technique substantially improves on this baseline, yielding a new state of the art. Our findings and analysis affirm the importance of structural biases, even in models that learn from large amounts of data.
\end{abstract}

\ignore{
\begin{abstract}
As language exhibits hierarchical structure, we address the problem of designing language models that benefit from hierarchical bias and can capture complex syntactic dependencies to a large extent, albeit without introducing much overhead in terms of scalability. To this end, we employ knowledge distillation (KD) as a means of injecting hierarchical bias from a slow syntactic language model that leverages syntactic information, to a more scalable student LSTM that operates sequentially. While we find that LSTM language models are able to achieve better syntactic generalisation than previously thought, our proposed combination of hierarchical bias and scalability improves over this strong baseline and yields a new state of the art on targeted syntactic evaluations. Analysis suggests that, compared to LSTM language models, our approach has better sample complexity for capturing complex syntactic constructions and encodes hierarchical information to a large extent, despite lacking direct access to syntactic annotation.  
\end{abstract}
}

\section{Introduction}
\ignore{
Despite the widespread success of recurrent architectures such as LSTMs \cite{hochreiter_97} in language processing tasks, recent work has highlighted the shortcomings of LSTM language models (LMs) in capturing complex syntactic dependencies \cite{marvin_2018}. 

To address this problem, injecting syntactic information into language models at training time---with the goal of improving structural generalisation---remains an active area of research. Existing methods largely fall into two categories: (i) \emph{multi-task learning} with syntactic objectives as auxiliary losses \cite{nadejde_2017,enguehard_2017,marvin_2018} and (ii) \emph{syntactic language models} that jointly generate surface strings ($\boldsymbol{x}$) and tree structures ($\boldsymbol{y}$) \cite[\emph{inter alia}]{baker-79,charniak_2001}. }

Language models (LMs) based on sequential LSTMs~\citep{hochreiter_97} have numerous practical applications, but it has also been shown that they do not always develop accurate syntactic generalisations~\citep{marvin_2018}. Thus, one strategy for improving LSTMs is to change their biases to facilitate more linguistically valid generalisations. 

This paper introduces a scalable method for introducing syntactic biases to LSTMs (and indeed, to any left-to-right language model trained with a cross-entropy objective) by distilling knowledge~\citep{bucila:2006,dark_knowledge} from recurrent neural network grammars \cite[RNNGs]{rnng}. RNNGs have been shown to 
successfully capture non-local syntactic dependencies 
\cite{kuncoro_2018}, achieve excellent parsing performance \cite{kuncoro-2017,fried_17}, and correlate well with encephalography signals \cite{hale_2018}. Unfortunately, these benefits come at the expense of \emph{scalability}, since the hierarchical constituent composition process (\S\ref{sec:rnng}) within RNNGs means that the structure of the computation graph for a sentence varies according to its tree structure. Even with the help of automatic dynamic batching \cite{dynet,neubig_2017}, RNNGs can be ten times slower to train than a comparable LSTM as they benefit less from specialised hardware like GPUs.
As such, RNNGs are an impractical alternative to computationally convenient architectures that are used to build language models from massive corpora~\citep{peters_2018,howard_2018,devlin_2019,radford_2019}.


As RNNGs are hard to scale, we instead use the predictions of an RNNG \emph{teacher model} trained on a small training set, to guide the learning of syntactic structure in a sequential LSTM \emph{student model}, which is trained on the training set in its entirety. We denote the resulting lanugage model (i.e., the student LSTM) as a \emph{distilled syntax-aware} LSTM LM (\textbf{DSA-LSTM}). Intuitively, the RNNG teacher is an expert on syntactic generalisation, although it lacks the opportunity to learn the relevant semantic and common-sense knowledge from a large training corpus. By learning from both, the DSA-LSTM therefore learns from a signal that is informative for syntactic generalisation, but without sacrificing the semantic richness contained in a large corpus. 

Since the DSA-LSTM differs from a conventional LSTM only in its training loss, it has the same hardware-friendly computational structure as a conventional LSTM, and is therefore much faster to train. On targeted syntactic evaluations, it achieves better accuracy than: (i) a strong LSTM LM which, through careful hyperparameter tuning, performs much better than previously thought (\S\ref{sec:replication}); (ii) the teacher RNNG that exploits a hierarchical inductive bias but lacks scalability (\S\ref{sec:rnng}); and (iii) a born-again network \cite{furlanello_18} that similarly learns from KD, albeit without a hierarchical bias from the teacher. We analyse the DSA-LSTM's internal representation through the syntactic probe \cite{shi_16,adi_2017} of \newcite{blevins_2018}, and find that the learned representations encode hierarchical information to a large extent, despite the DSA-LSTM lacking direct access to syntactic annotation.

While not directly comparable, on subject-verb agreement both the teacher RNNG and student DSA-LSTM outperform BERT \cite{devlin_2019,goldberg_2019}, which benefits from bidirectional information and is trained on 30 times as much data. Altogether, these findings suggest that structural biases continue to play an important role, even at massive data scales, in improving the linguistic competence of LMs.

\section{Replication of Targeted Syntactic Evaluations of LSTM LMs}\label{sec:replication}

In this section, we replicate the targeted syntactic evaluations reported by \newcite{marvin_2018}, which 
assess LMs' ability to assign higher probability in grammatical/ungrammatical minimal pairs within a variety of complex syntactic structures. This will serve as our primary evaluation instrument in this paper.


The following example illustrates the \emph{subject-verb agreement across an object relative clause (no complementiser)} construction:

\vspace{-2mm}
\begin{itemizesquish}
\item The farmer the parents love \underline{swims}/$^{*}$\underline{swim}.
\end{itemizesquish}
\vspace{-2mm}
\noindent An LM succeeds on each example iff it assigns a higher probability to the grammatical sentence.
\newcite{marvin_2018} report that LSTMs, even with multi-task syntactic supervision, on aggregate still lag far behind human performance.

\vspace{-2mm}
\paragraph{Experimental settings.} Following \newcite{marvin_2018}, we use LSTMs with 650 hidden units trained on the Wikipedia corpus of \newcite{gulordava_18}. Hyperparameters are optimised based on a grid search and can be found in the Appendix. As the targeted syntactic evaluations are 
based on 
individual sentences, our LSTM models each sentence separately.\footnote{By modelling each sentence separately, our setup is consistent with that of \newcite{marvin_2018} but differs from those with cross-sentential context \cite{mikolov:2010}.} 
\vspace{-2mm}
\paragraph{Discussion.} We present our findings in Table \ref{tab:replication} ({\bf ``Ours''}); for all our models we report mean and standard deviation of 10 identical models from different random seeds. Our LSTM LM achieves much better perplexity than the LSTM LM (32\% ppl.\ reduction) and even the multi-task LSTM (12\% reduction) of \newcite{marvin_2018}. As our LSTM has the same number of hidden units, we attribute this gap to differences in optimisation and codebases. 
On aggregate, our LSTM LM outperforms the LSTM multi-task model from \newcite{marvin_2018} that exploits explicit CCG annotations, and is able to match or exceed human performance on 7 out of all 15 constructions, thus confirming earlier findings that neural language models are able to acquire complex syntactic generalisation without explicit syntactic supervision \cite{gulordava_18,goldberg_2019}. 

Despite the small variance in perplexity (stdev 0.16 ppl.), the trained LMs exhibit large variance in accuracy for some constructions (up to stdev 0.12 for NPI across a relative clause). This observation is consistent with earlier findings that models with similar perplexity may exhibit different patterns of syntactic generalisation \cite{kuncoro_2018,tran_2018}, and serves as a caution against reporting results based on single runs.

\begin{table*}[t]
      \centering
      \resizebox{\textwidth}{!}{%
       \begin{tabular}{l|r|r|r||r|r||r}
       &  \multicolumn{2}{c|}{\textbf{Marvin \& Linzen models}} &  \multicolumn{1}{c||}{\textbf{Ours}} & \multicolumn{2}{c||}{\textbf{Ours (small training)}} & \\ \hline
		&   \textbf{M\&L-LSTM} & \textbf{M\&L-Multi}   & \textbf{Our LSTM}  & \textbf{Small LSTM}$^{\dagger}$ & \textbf{RNNG}$^{\dagger}$ & \textbf{Humans}   \\
		\hline\hline
		\newcite{gulordava_18} \textbf{test perplexity} &  78.65 & 61.10 & \textbf{53.73} ($\pm$0.16)   & 94.54 ($\pm$0.21) & \textbf{92.30} ($\pm$0.27) & N/A \\ \hline \hline
	    \multicolumn{7}{c}{SUBJECT-VERB AGREEMENT}  \\ \hline
		Simple &  0.94 & \textbf{1.00} & \textbf{1.00} ($\pm$0.00)  & 0.89  ($\pm$0.03) & \textbf{0.99} ($\pm$0.01) & 0.96 \\ 
        In a sentential complement & \textbf{0.99} & 0.93  & 0.97 ($\pm$0.02) &  0.89 ($\pm$0.01) & \textbf{0.93} ($\pm$0.02) & 0.93 \\ 
        Short VP coordination & 0.90 & 0.90 & \textbf{0.96} ($\pm$0.02)  & 0.90 ($\pm$0.03) & \textbf{0.96} ($\pm$0.02) & 0.94 \\ 
        Long VP coordination  & 0.61 & 0.81 & \textbf{0.82} ($\pm$0.05)  & 0.78 ($\pm$0.03) & \textbf{0.94} ($\pm$0.03) & 0.82 \\ 
        Across a prepositional phrase  & 0.57 & 0.69 & \textbf{0.89} ($\pm$0.02)   & 0.83 ($\pm$0.02) & \textbf{0.95} ($\pm$0.01) & 0.85 \\ 
        Across a subject relative clause  & 0.56 & 0.74 & \textbf{0.87} ($\pm$0.02)   & 0.81 ($\pm$0.04) & \textbf{0.95} ($\pm$0.03) & 0.88 \\
        Across an object relative clause  & 0.50 & 0.57 & \textbf{0.77} ($\pm$0.11)   & 0.54 ($\pm$0.08) & \textbf{0.95} ($\pm$0.03) & 0.85 \\
        Across an object relative clause (no \emph{that}) & 0.52 & 0.52 & \textbf{0.70} ($\pm$0.05)  & 0.55 ($\pm$0.07) & \textbf{0.93} ($\pm$0.02) & 0.82 \\
        In an object relative clause  & 0.84 & 0.89 & \textbf{0.90} ($\pm$0.03)  & 0.79 ($\pm$0.05) & \textbf{0.96}  ($\pm$0.01) & 0.78 \\
        In an object relative clause (no \emph{that})  & 0.71 & 0.81 & \textbf{0.86} ($\pm$0.05)   & 0.72 ($\pm$0.03) & \textbf{0.96} ($\pm$0.02) & 0.79 \\ \hline 
        \textbf{Average of subject-verb agreement}  & 0.71 & 0.79 & \textbf{0.87} ($\pm$0.02)   & 0.77 ($\pm$0.02) & \textbf{0.95} ($\pm$0.01) & 0.86  \\ \hline \hline
        \multicolumn{7}{c}{REFLEXIVE ANAPHORA}  \\ \hline
        Simple  & 0.83 & 0.86 & \textbf{0.91} ($\pm$0.01)   & \textbf{0.93} ($\pm$0.01) & 0.83 ($\pm$0.02) & 0.96 \\ 
       In a sentential complement  & \textbf{0.86} & 0.83 & 0.81 ($\pm$0.02)   & \textbf{0.77} ($\pm$0.03) & 0.46 ($\pm$0.05) & 0.91  \\ 
       Across a relative clause  & 0.55 & 0.56 & \textbf{0.64} ($\pm$0.02)   & 0.63 ($\pm$0.02) & \textbf{0.82} ($\pm$0.02) & 0.87  \\ \hline 
        \textbf{Average of reflexive anaphora}   & 0.75 & 0.75 & \textbf{0.79} ($\pm$0.01)     & \textbf{0.78} ($\pm$0.01) & 0.70 ($\pm$0.02) & 0.91 \\ \hline \hline
       \multicolumn{7}{c}{NEGATIVE POLARITY ITEMS}  \\ \hline
       Simple  & 0.40 & 0.48 & \textbf{0.96} ($\pm$0.04)   & \textbf{0.93} ($\pm$0.06) & 0.28 ($\pm$0.05) & 0.98 \\ 
       Across a relative clause  & 0.41 &  0.73 & \textbf{0.75} ($\pm$0.12)  & \textbf{0.82} ($\pm$0.09) & 0.78 ($\pm$0.06)  &  0.81  \\ \hline
        \textbf{Average of negative polarity items}   & 0.41 & 0.61 & \textbf{0.86} ($\pm$0.06)   & \textbf{0.88} ($\pm$0.05) & 0.53 ($\pm$0.04) & 0.90 \\ \hline \hline
       \textbf{Average of all constructions}  & 0.68 & 0.75 & \textbf{0.85} ($\pm$0.02)  &  0.79 ($\pm$0.02) & \textbf{0.85} ($\pm$0.02) &  0.88  
		\end{tabular}}
		\caption{Replication of \newcite{marvin_2018} results. M\&L-Multi is the \newcite{marvin_2018} LSTM trained on LM and CCG supertagging \cite{bangalore_1999,clark-curran-2007-wide} losses with an interpolation factor of 0.5. We report our LSTM LM, small LSTM$^{\dagger}$, and RNNG$^{\dagger}$ performance ($^{\dagger}$smaller training data; \S\ref{sec:rnng}) in the format of \emph{mean ($\pm$standard deviation)} of 10 identical models from different seeds. Results in bold denote the best among models trained on similar amounts of training data.
		}
        \label{tab:replication}
        \vspace{-4mm}
\end{table*}
\vspace{-2mm}
\section{Syntactic Evaluation with RNNG}\label{sec:rnng}
\vspace{-2mm}
\emph{To what extent can a model that leverages syntactic bias and annotation do well on targeted syntactic evaluations, even when trained on less data?}

Here we briefly describe and assess the performance of the stack-only RNNG \cite{kuncoro-2017} that we use as the teacher. Our choice of RNNG is motivated by its excellent number agreement performance on the \newcite{linzen-2016} dataset,\footnote{While BERT \cite{devlin_2019} achieves even better number agreement performance \cite{goldberg_2019}, the results are not directly comparable since BERT operates non-incrementally and was trained on 500 times as much data. The current state of the art among models trained on the \newcite{linzen-2016} training set is the adaptive universal transformer model \cite{dehghani_19}.} achieving 92.9\% for four attractors under purely incremental decoding \cite{kuncoro_2018}. 

\subsection{Recurrent Neural Network Grammars}

An RNNG defines the joint probability of surface string $\boldsymbol{x}$ and phrase-structure tree $\boldsymbol{y}$, denoted as $t(\boldsymbol{x}, \boldsymbol{y})$. The model generates phrase-structure trees in a top-down, left-to-right manner through a series of action sequences in a process reminiscent of shift-reduce parsing. At any given state, the decision over which action to take is parameterised by a stack LSTM~\citep{stack_lstm} encoding partially-completed constituents. Let $\mathbf{h}_t$ be the stack LSTM hidden state at time $t$. The next action $a_t \in \{ \textsc{gen}, \textsc{nt}, \textsc{reduce}\}$ is sampled according to a categorical distribution defined by an affine transformation and a softmax:
\vspace{-2mm}
\begin{equation*}
    a_t \sim \softmax(\mathbf{W_a}\mathbf{h}_t + \mathbf{b_a}).
\end{equation*}
\vspace{-9mm}

 \begin{itemizesquish}
 \item If $a_t \in \{ \textsc{gen}, \textsc{nt}\}$, the model samples a terminal $x$ or a non-terminal $n$ from each respective categorical distribution as the next input:
 \end{itemizesquish}
 \vspace{-6mm}
\begin{align*}
    x \sim \softmax(\mathbf{W_x}\mathbf{h}_t + \mathbf{b_x}), \\
    n \sim \softmax(\mathbf{W_n}\mathbf{h}_t + \mathbf{b_n}).
\end{align*}
\vspace{-9mm}
\begin{itemizesquish}
\item If $a_t = \textsc{reduce}$, the topmost stack elements going back to the last incomplete non-terminal are popped, and a \emph{composition function} (here a bidirectional LSTM) is executed to represent the completed phrase on the stack. This recursive composition function constitutes a primary difference with the syntactic LM of \newcite{choe:2016} that operates sequentially, and has been found to be crucial for achieving good number agreement \cite{kuncoro_2018} and correlation with brain signals \cite{hale_2018}.
 \end{itemizesquish}
  \vspace{-3mm}
The stack LSTM, composition function, lookup embeddings, and pairs of affine transformation weights and biases $\{\mathbf{W}, \mathbf{b}\}$ are model parameters. 

\subsection{Experiments}
Here we outline the experimental settings and present our RNNG findings. 
\vspace{-2mm}
\paragraph{Experimental settings.} We implement the RNNG with DyNet and enable autobatching on GPU. Predicted phrase-structure trees for the training and validation sets of the \newcite{gulordava_18} Wikipedia dataset are obtained with a pre-trained Berkeley parser \cite{petrov:2007}. Since training the RNNG on the full training set with the same number of hidden units as the LSTM would take more than a month,\footnote{We tested the speed of RNNGs and LSTMs with similar capacity (40 million parameters) on DyNet. Both models ran on a single Quadro P4000 GPU with automatic batching turned on and a batch size of 20 sentences.} we train the RNNG on $\sim20\%$ of the training set (600,000 sentences), and use a smaller hidden state size of 256 (vs. 650 for the full LSTM). As the dataset is pre-processed, we select this subset such that all word types occur at least once in this smaller training set. 
\vspace{-2mm}
\paragraph{Incremental decoding and marginal probability.} To preserve incrementality constraints, at test time we use a word-synchronised beam search \cite{fried_17} with fast-tracking \cite{stern_17}, using word and action beam sizes of $k=50$ and $k \times 10=500$, respectively. As exact inference of $t(\boldsymbol{x})$ is intractable, we evaluate with a lower bound of the marginal probability by summing over the top $k$ hypotheses $\boldsymbol{y}^{b(\boldsymbol{x})}_1,  \ldots, \boldsymbol{y}^{b(\boldsymbol{x})}_k$ on the beam $b(\boldsymbol{x})$ once parsing finishes:
\vspace{-2mm}
\begin{align*}
   t(\boldsymbol{x}) =  \sum_{\boldsymbol{y'} \in \mathcal{T}(\boldsymbol{x})} t(\boldsymbol{x}, \boldsymbol{y'})
    \geq \sum_{i=1}^k t(\boldsymbol{x}, \boldsymbol{y}^{b(\boldsymbol{x})}_i),
\end{align*}
where $\mathcal{T}(\boldsymbol{x})$ denotes the set of all possible phrase-structure trees for a sentence $\boldsymbol{x}$. On targeted syntactic evaluations, the model succeeds iff $\log t(\boldsymbol{x}_{\text{correct}}) > \log t(\boldsymbol{x}_{\text{incorrect}})$. 
\vspace{-2mm}
\paragraph{Discussion.} We present the results in Table \ref{tab:replication} (sixth column: \textbf{``RNNG''}), and compare with LSTMs trained on: (i) the full dataset (fourth column: \textbf{``Our LSTM''}), and (ii) the same (smaller) training set as the RNNG (fifth column: \textbf{``Small LSTM''}). 
Our findings clearly reaffirm the benefits of \emph{both} hierarchical bias and data scale. In terms of hierarchical bias, an RNNG that leverages syntactic annotations and explicit composition operators outperforms a comparable small LSTM on 11 out of 15 constructions, and on aggregate improves accuracy on targeted syntactic evaluations from 79\% to 85\% (29\% error reduction), thus matching the aggregate performance of the full LSTM trained on 5 times as much data, although we remark that their success and failure modes appear to be different. 

In terms of data scale, the LSTM LM trained on the full training set substantially outperforms the LSTM trained on the smaller training set. In particular, the performance difference between the small and full LSTMs sheds light on which constructions are sensitive to variations in the amount of data. For instance, agreement across an object relative clause exhibits large variations across the two training regimes (77\% to 54\%), suggesting that LSTMs require a large amount of data to learn these constructions well. Our finding on the 
importance of data scale for LM training is consistent with the success of recent LM pre-training approaches \cite[\emph{inter alia}]{peters_2018,devlin_2019} and earlier work on noisy channel models for tasks such as machine translation and speech recognition \cite[\emph{inter alia}]{jelinek_1998,rosenfeld_2000,koehn_2010}. 

Despite its smaller training set, the RNNG performs extremely well on subject-verb agreement, substantially outperforming both the full LSTM and a pre-trained BERT \cite[Table~\ref{tab:dsa-lstm}]{devlin_2019} trained on 150 times as much data, although it still lags behind the full LSTM on reflexive anaphora and NPI. 
\vspace{-2mm}
\section{Syntax-Aware Language Model}\label{sec:synaware}
\emph{Given the trade-off between hierarchical operations and scalability, how can we design LMs that can better capture complex syntactic dependencies \textbf{and} be easily scalable at the same time?}
\vspace{-2mm}
\subsection{Knowledge Distillation (KD)}
The goal of KD is to find a set of student model parameters $\hat{\theta}_{\text{KD}}$ that would minimise the Kullback–Leibler (KL) divergence between the teacher RNNG's marginal probability $t(\boldsymbol{x}) = \sum_{\boldsymbol{y'} \in \mathcal{T}(\boldsymbol{x})} t(\boldsymbol{x}, \boldsymbol{y'})$ and the LSTM student $q_{\theta}(\boldsymbol{x})$. Expanding the KL term and removing terms that do not depend on $\theta$ yields:
\vspace{-2mm}
\begin{align}
\hat{\theta}_{\text{KD}} &=\argmin_{\theta} D_{\mathrm{KL}} \left(t(\boldsymbol{x}) \,\,|| \,\, q_{\theta}(\boldsymbol{x}) \right), \\
      &= \argmin_{\theta} -\sum_{\boldsymbol{x} \in \Sigma^{\mathbf{*}}} t(\boldsymbol{x}) \log q_{\theta}(\boldsymbol{x}), \label{eq:kd_intractable} \\ 
      &= \argmin_{\theta} - \E_{\boldsymbol{x} \sim t(\boldsymbol{x})}  \log q_{\theta}(\boldsymbol{x}), \label{eq:kd_intractable_exp}
\end{align}
where $\Sigma$ denotes the set of all word types in the vocabulary, and $\Sigma^{\mathbf{*}}$ the set of all possible sentences. As Eq.~\ref{eq:kd_intractable} involves an intractable summation over the set of all possible sentences, one alternative is to approximate this expectation with Monte Carlo sampling to obtain $K$ sentences $D' = \{\boldsymbol{x}'^{(1)}, \ldots, \boldsymbol{x}'^{(K)}\}$ from $t(\boldsymbol{x})$,\footnote{While an RNNG estimates $t(\boldsymbol{x}, \boldsymbol{y})$, a simple way of sampling surface strings $\boldsymbol{x}$ from the RNNG is to sample pairs of $(\boldsymbol{x}^{(k)}, \boldsymbol{y}^{(k)}) \sim t(\boldsymbol{x}, \boldsymbol{y})$ and ignore all non-terminals $\boldsymbol{y}^{(k)}$.} and train a student LSTM LM on these sampled sentences as opposed to ground-truth LM data:
\vspace{-3mm}
\begin{align*}
     \E_{\boldsymbol{x} \sim t(\boldsymbol{x})}  \log q_{\theta}(\boldsymbol{x})
     \approx \dfrac{1}{K} \sum_{\boldsymbol{x}' \in D'} \sum_{j=1}^{|\boldsymbol{x}'|} \log q_{\theta}(x'_j | \boldsymbol{x}'_{<j}), \label{eq:sampling_rnng}
\end{align*}
although our preliminary experiments suggest that this procedure performs poorly due to high variance.\footnote{This procedure of training a student LSTM LM on string samples from the RNNG with $K \approx 3,000,000$ yields a high validation perplexity of above 1,000, due to the enormity of the sample space and the use of discrete samples.} We instead approximate Eq.~\ref{eq:kd_intractable_exp} by minimising the KL at the \emph{local word-level}: 
\vspace{-2mm}
\begin{align*}
    &\E_{\boldsymbol{x \sim t(\boldsymbol{x})}}  \log q_{\theta}(\boldsymbol{x}) \approx\\
    & \E_{\boldsymbol{x^{*}} \sim p^{\mathbf{*}}(\boldsymbol{x})}\,  \sum_{j=1}^{|\boldsymbol{x^{\mathbf{*}}}|}  D_{\mathrm{KL}} \left( t(w \mid \boldsymbol{x^{\mathbf{*}}}_{<j}) \,\,|| \,\, q_{\theta}(w \mid \boldsymbol{x^{\mathbf{*}}}_{<j}) \right),
\end{align*}
where $\boldsymbol{x^{\mathbf{*}}}$ is sampled from the empirical distribution $p^{\mathbf{*}}(\boldsymbol{x})$, rather than from the teacher RNNG. Here $t(w \mid \boldsymbol{x^{\mathbf{*}}}_{<j})$ and $q_{\theta}(w \mid \boldsymbol{x^{\mathbf{*}}}_{<j})$ respectively parameterise the (marginal) probability of generating the next-word continuation $w \in \Sigma$, given the ``ground-truth'' conditioning context $\boldsymbol{x^{\mathbf{*}}}_{<j}$, under the teacher and student models.

\begin{figure*}[t]
    \centering
    \includegraphics[scale=0.48]{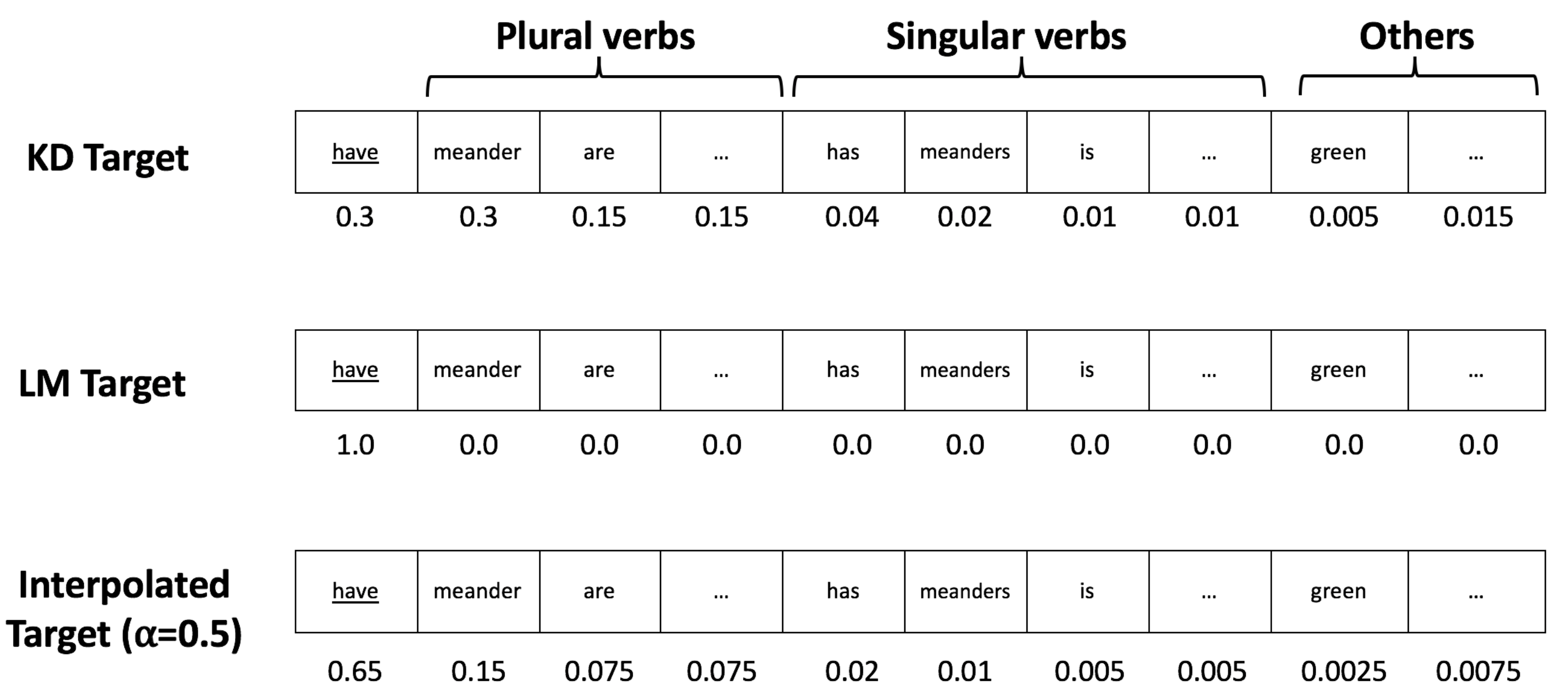}
    \vspace{-2mm}
    \caption{Example of the KD target (top), the standard LM target (middle), and the interpolated target used to train the DSA-LSTM (bottom) with $\alpha=0.5$, for a prefix (showing only the terminals) \emph{Parts of the river valley}, where the correct continuation is \emph{\underline{have}} due to the plural subject \emph{parts}.}
    \label{fig:distillation}
     \vspace{-4mm}
\end{figure*}

For a dataset of sentences $D = \{\boldsymbol{x^{\mathbf{*}}}^{(1)}, \ldots, \boldsymbol{x^{\mathbf{*}}}^{(|D|)} \}$ characterising the empirical distribution $p^{\mathbf{*}}(\boldsymbol{x^{\mathbf{*}}})=\frac{1}{|D|}$ when $\boldsymbol{x^{\mathbf{*}}} \in D$ (i.i.d. assumption), the word-level objective is:
\vspace{-2mm}
\begin{align*}
  &\hat{\theta}_{\text{KD}} \approx \argmin_{\theta} - \dfrac{1}{|D|} \sum_{\boldsymbol{x^{\mathbf{*}}} \in D} \ell_{\text{KD}}(\boldsymbol{x^{\mathbf{*}}}; \theta), \\
  &\ell_{\text{KD}}(\boldsymbol{x^{\mathbf{*}}}; \theta) =  \sum_{j=1}^{|\boldsymbol{x^{\mathbf{*}}}|} \sum_{w \in \Sigma} t(w \mid \boldsymbol{x^{\mathbf{*}}}_{<j}) \log q_{\theta}(w \mid \boldsymbol{x^{\mathbf{*}}}_{<j}).
\end{align*}
In earlier work, this local word-level approximation to the KD objective for sequence models has been shown to work surprisingly well in the case of neural machine translation\footnote{While \newcite{seq-distillation} proposed a technique for sequence-level KD for machine translation through beam search, the same technique is not directly applicable to LM, which is an unconditional language generation problem.} \cite{seq-distillation} and language modelling \cite[Born-Again Networks]{furlanello_18}.
\vspace{-2mm}
\paragraph{Interpolation.} As the teacher RNNG is trained on a smaller training set, the DSA-LSTM should not only aim to emulate the RNNG's predictions and risk being upper-bounded by the teacher's performance, but also learn from the correct next word $x^{*}_j$ to fully exploit scalability.\footnote{Recall that $\ell_{\text{KD}}(\boldsymbol{x}; \theta)$ does not depend on the true next word $x^{*}_j$.} We thus interpolate the distillation (left) and LM (right) losses:
\vspace{-5mm}
\begin{align*}
    &\hat{\theta}_{\alpha\text{-int}} = \argmin_{\theta} -\dfrac{1}{|D|} \sum_{\boldsymbol{x^{\mathbf{*}}} \in D} \\
    & \left[ \alpha \ell_{\text{KD}}(\boldsymbol{x^{\mathbf{*}}}; \theta) + (1 - \alpha) \sum_{j=1}^{|\boldsymbol{x^{\mathbf{*}}}|} \log q_{\theta}(x^{*}_j \mid \boldsymbol{x^{\mathbf{*}}}_{<j}) \right],
\end{align*}
where $\alpha$ is the interpolation coefficient. We illustrate the effect of this interpolation in Fig.~\ref{fig:distillation}.

Furthermore, computing $\ell_{\text{KD}}(\boldsymbol{x^{\mathbf{*}}}; \theta)$ requires the RNNG's estimate of $t(w\mid\boldsymbol{x^{\mathbf{*}}}_{<j})$, which necessitates an expensive marginalisation over all tree prefixes that generate $w$ conditional on $\boldsymbol{x^{\mathbf{*}}}_{<j}$. For efficiency, we approximate this using the one-best predicted tree from a pre-trained Berkeley parser,\footnote{We use the same pre-trained Berkeley parser to obtain training and validation trees in \S\ref{sec:rnng}.} denoted as $\hat{\boldsymbol{y}}^{\text{berk}}(\boldsymbol{x^{\mathbf{*}}})$, as follows:
\vspace{-2mm}
\begin{align*}
    t(w \mid \boldsymbol{x^{\mathbf{*}}}_{<j}) &\approx t(w \mid \boldsymbol{x^{\mathbf{*}}}_{<j},\,\, \hat{\boldsymbol{y}}^{\text{berk}}_{<j}(\boldsymbol{x^{\mathbf{*}}})),
\end{align*}
where $\hat{\boldsymbol{y}}^{\text{berk}}_{<j}(\boldsymbol{x^{\mathbf{*}}})$ are all the non-terminals in $\hat{\boldsymbol{y}}^{\text{berk}}(\boldsymbol{x^{\mathbf{*}}})$ that occur before $x^{*}_j$. In other words, we first parse the sentence with a Berkeley parser, and use the resulting tree prefix as conditioning context to compute the probability of generating $w \in \Sigma$ under the RNNG. While this means that the teacher's predictions are not derived from a purely incremental process,\footnote{The resulting syntactic prefix $\hat{\boldsymbol{y}}^{\text{berk}}_{<j}(\mathbf{x})$ for approximating $t(w \mid \boldsymbol{x^{\mathbf{*}}}_{<j})$ under the RNNG is obtained from a Berkeley parser that has access to yet unseen words $\boldsymbol{x}_{>j}$.} the student DSA-LSTM still operates strictly incrementally. This interpolated objective is similar to label smoothing \cite{szegedy_2016,pereyra_2017}, with the softmax distribution of the RNNG as the smoothing factor as opposed to the uniform distribution.
\vspace{-2mm}
\paragraph{Intuition.} In Fig.~\ref{fig:distillation}, we provide an intuition about why the interpolation of the distillation and LM losses could inject hierarchical bias into a sequential model. We consider the interpolated target with $\alpha=0.5$ for a prefix (suppressing non-terminals) \emph{Parts of the river valley}, where the correct continuation is \emph{have} since the agreement controller \emph{parts} is plural. The standard LM loss is zero only when all word types other than the correct one are assigned zero probability mass, and it is only in expectation (across training contexts) that syntactic regularities are inferred. In contrast, the interpolated target assigns a minimum probability of 0.5 to the correct label, but crucially contains additional information about the \emph{plausibility} of every alternative based on the teacher RNNG's predictions. Under this objective, the plural verbs \emph{are} and \emph{meander} are assigned relatively high probability mass since they fit both the syntactic and semantic constraints (e.g. Parts of the river valley often meander), while the set of singular verbs \emph{has}, \emph{meanders}, and \emph{is} are assigned much lower probability mass since they are syntactically illicit. Thus, as long as the RNNG makes the accurate structural generalisations (and we have shown that it largely does in \S\ref{sec:rnng}), every training instance provides the student LSTM with a wealth of information about all the possible legitimate continuations according to the predictions of the hierarchical teacher, 
thereby making it easier for the student to learn the appropriate hierarchical constraints and generalisations.

\vspace{-2.5mm}
\paragraph{Differences with other KD work.} Our approach departs from the predominant view of distillation primarily as a means of \emph{compressing} knowledge from a bigger teacher or an ensemble to a compact student \cite[\emph{inter alia}]{ba_et_al_14,seq-distillation,liu_2018} in two important ways. First, here the teacher and student models are different in character, and not just in size: we transfer knowledge from a teacher that models the joint probability of strings and phrase-structure trees through hierarchical operations, to a student that only models surface strings through sequential operations. This setup presents an interesting dynamic since the DSA-LSTM has to mimic the predictions of the RNNG, which conditions on syntactic annotation to guide hierarchical operations, even though the DSA-LSTM itself has no direct access to any syntactic annotations at all. 

Second, distillation thus far has mostly been applied in settings where the teacher and student models are trained on the same data. For scalability reasons, we train the RNNG on a subset of the data, and obtain its soft predictions on the rest. We hypothesise that the predictions of the hierarchical teacher---although they come from a model trained on a smaller dataset---can nevertheless encourage the LSTM to develop structurally sensitive representations of the larger dataset it observes.
\vspace{-2.5mm} 
\paragraph{Born-Again Networks (BA).} In practice, the interpolated distillation objective above can be applied to any teacher and student models. Recently, \newcite{furlanello_18} surprisingly finds perplexity improvement in a \emph{born-again} setup that trains an LSTM LM on the gold data, and then uses the resulting model as a teacher to a student LSTM that shares the same architecture as the teacher. To better understand the importance of learning from a hierarchical teacher (which is not the case in a BA-LSTM since the teacher model is also sequential), we present experiments comparing the DSA-LSTM with a BA-LSTM. 

\subsection{Experiments}
Here we describe our experimental settings and present our findings.
\vspace{-2mm}
\paragraph{Computational challenge.}
The KD loss necessitates computing the teacher RNNG's predictive softmax distribution for each token in the training set, but pre-computing these for the \newcite{gulordava_18} training set leads to a prohibitive memory footprint.\footnote{Pre-computing the RNNG's predictions necessitates storing $N \times |\Sigma|$ numbers, where $N$ is the number of tokens. For the \newcite{gulordava_18} training set ($\sim$80M tokens), this requires storing 4 trillion floating points, or 25 terabytes.} 
To save space, we instead pre-compute the teacher RNNG's hidden state $\mathbf{h}_t \in \mathcal{R}^M$ for every token $x_t$ in the training set ($M \ll |\Sigma|$), and compute the teacher's softmax on-the-fly with an affine transformation and a softmax, which presents minimal computational overhead.
\vspace{-2.5mm}
\paragraph{Experimental settings.} The DSA-LSTM has an identical architecture to the LSTM LM (\S\ref{sec:replication}), although the learning rate is optimised independently (Appendix). 
 We select the final model based on validation LM perplexity, with targeted syntactic evaluations only applied at test time.
\paragraph{Training speed.} Since the DSA-LSTM operates sequentially, it is amenable to batching operations and is five times faster to train than a comparable RNNG. Despite this significant speed-up, training the DSA-LSTM in our basic implementation is still half as fast as the standard LM objective. We attribute this difference to the additional computational overhead associated with the distillation objective, such as I/O operations and computing the cross-entropy between the teacher and student models for the entire vocabulary. These operations, however, only apply at training time; at test time there is no overhead of inferring $q_{\hat{\theta}_{\alpha\text{-int}}}(\boldsymbol{x})$ under the DSA-LSTM. 
\paragraph{Baselines.} The DSA-LSTM benefits from three main components: (i) a KD objective, which in itself has been shown to be a good regulariser \cite{furlanello_18}, (ii) the scalability of the sequential architecture, and (iii) a hierarchical bias, which here comes from the teacher RNNG. To understand the benefit of each component, we compare DSA-LSTM with these baselines:
\begin{itemizesquish}
\item a strong LSTM LM (\S\ref{sec:replication}) that is scalable but lacks a hierarchical bias (\textbf{``Full LSTM''});
\item the teacher RNNG trained on a 20\% subset of the training set (\S\ref{sec:rnng}), which benefits from a hierarchical bias but lacks scalability (\textbf{``RNNG''});
\item a DSA-LSTM trained on the same smaller subset as the teacher RNNG (\textbf{``S-DSA-LSTM''}). This baseline isolates the importance of scalability, since it still benefits from a KD objective and a hierarchical bias from the teacher RNNG;
\item a born-again LSTM that benefits from KD and scalability, though it lacks a hierarchical bias due to the sequential teacher (\textbf{``BA-LSTM''}).
\end{itemizesquish}
\begin{table*}[!htb]
      \centering
      \resizebox{\textwidth}{!}{%
       \begin{tabular}{l|r|r|r||r|r|r||r||r}
       & \multicolumn{3}{c||}{\textbf{Small Training Set}} & \multicolumn{3}{c||}{\textbf{Full Training Set}} \\ \hline
		& \textbf{Small LSTM}$^{\dagger}$ & \textbf{S-DSA-LSTM}$^{\dagger}$ &  \textbf{RNNG}$^{\dagger}$ & \textbf{Full LSTM} & \textbf{BA-LSTM} & \textbf{DSA-LSTM} &\textbf{BERT} & \textbf{Humans}  \\
		\hline\hline
		\newcite{gulordava_18} \textbf{test ppl.} & 94.54 & 93.95  & \textbf{92.30} & \textbf{53.73}  &  54.64 &  56.74 & N/A & N/A \\ \hline \hline
	    \multicolumn{9}{c}{SUBJECT-VERB AGREEMENT}  \\ \hline
		Simple & 0.89 & 0.96  & \textbf{0.99} & \textbf{1.00}   & \textbf{1.00}  & \textbf{1.00}  & 1.00 & 0.96 \\ 
        In a sentential complement & 0.89 & \textbf{0.98}   & 0.93 & 0.97  & \textbf{0.98}  & \textbf{0.98}  & 0.83 & 0.93 \\ 
        Short VP coordination & 0.90 &  0.88  & \textbf{0.96} & 0.96 & 0.95   & \textbf{0.99}    & 0.89 & 0.94 \\ 
        Long VP coordination & 0.78 & 0.74 & \textbf{0.94} & \textbf{0.82}  & 0.80  & 0.80    & 0.98 & 0.82 \\ 
        Across a prepositional phrase & 0.83 & 0.88   & \textbf{0.95} & 0.89  & 0.89   & \textbf{0.91}  & 0.85 & 0.85 \\ 
        Across a subject relative clause & 0.81 & 0.87 & \textbf{0.95} & 0.87  & 0.87  & \textbf{0.90}  & 0.84 & 0.88 \\
        Across an object relative clause & 0.54 & 0.69  & \textbf{0.95} & 0.77   &  0.81 & \textbf{0.84} & 0.89 & 0.85 \\
        Across an object relative clause (no \emph{that}) & 0.55 & 0.61  & \textbf{0.93} & 0.70   & 0.74   & \textbf{0.77} & 0.86 & 0.82 \\
        In an object relative clause & 0.79 & 0.87  & \textbf{0.96} & 0.90   & 0.91   & \textbf{0.92}  & 0.95 & 0.78 \\
        In an object relative clause (no \emph{that}) & 0.72 & 0.88  & \textbf{0.96} & 0.86   & 0.83  & \textbf{0.92} & 0.79 & 0.79 \\ \hline
         \textbf{Average of subject-verb agreement} & 0.77 & 0.84  & \textbf{0.95} & 0.87  & 0.88  & \textbf{0.90} & 0.89 & 0.86 \\ \hline \hline
        \multicolumn{9}{c}{REFLEXIVE ANAPHORA}  \\ \hline
        Simple & \textbf{0.93} & 0.90  & 0.83 & 0.91   & \textbf{0.92}   & 0.91  & 0.94 & 0.96\\ 
       In a sentential complement & 0.77 & \textbf{0.78} & 0.46 & 0.81   & 0.81  & \textbf{0.82}  & 0.89 & 0.91  \\ 
       Across a relative clause & 0.63 & 0.67   & \textbf{0.82} & 0.64   & 0.64  & \textbf{0.67}  & 0.80 & 0.87  \\ \hline
        \textbf{Average of reflexive anaphora} & \textbf{0.78} & \textbf{0.78}   & 0.70 & 0.79    & 0.79   & \textbf{0.80}  & 0.88 & 0.91 \\ \hline \hline
       \multicolumn{9}{c}{NEGATIVE POLARITY ITEMS}  \\ \hline
       Simple & 0.93 & 0.84  & 0.28 & 0.96   & \textbf{0.98}   & 0.94  & N/A & 0.98 \\ 
       Across a relative clause & 0.82 & 0.73   & 0.78 & 0.75 & 0.70   & \textbf{0.91}  & N/A &  0.81  \\ \hline
        \textbf{Average of negative polarity items} & \textbf{0.88} & 0.79  & 0.53 & 0.86  & 0.84   & \textbf{0.92}  & N/A & 0.90 \\ \hline \hline
       \textbf{Average of all constructions} & 0.79 & 0.82   & \textbf{0.85} & 0.85  & 0.86   & \textbf{0.89} & N/A &  0.88  
		\end{tabular}}
		\caption{Experimental findings of the \textbf{``DSA-LSTM''}. For each column, we report the mean of 10 identical models trained from different random seeds; standard deviation values are reported in the Appendix. \textbf{``S-DSA-LSTM''} indicates the DSA-LSTM trained on the smaller RNNG training set, while \textbf{``BA-LSTM''} is the born-again model where the teacher is the full LSTM LM. We also compare with the syntactic generalisation of \textbf{``BERT''} Base \cite{devlin_2019,goldberg_2019}, which is not strictly comparable since it is trained on 30 times as much data. $^{\dagger}$ indicates models trained on the smaller 20\% training set (\S\ref{sec:rnng}). Results in bold denote the best among those trained with the same amounts of data.}
        \label{tab:dsa-lstm}
        \vspace{-4mm}
\end{table*}
\vspace{-4mm}
\paragraph{Discussion.} To avoid clutter, for each model variant we present only the mean performance of 10 identical models from different random seeds; results with standard deviations are in the Appendix. We present our findings in Table~\ref{tab:dsa-lstm}, based on which we derive several observations.
\vspace{-2mm}
\begin{itemizesquish}
\item Of the three models trained on the small subset, the S-DSA-LSTM outperforms the small LSTM trained on standard LM objective, improving overall acccuracy from 0.79 to 0.82 (14\% error reduction), even though both models share the same architecture and training set size (i.e. only the training objective is different). On subject-verb agreement, the S-DSA-LSTM successfully narrows the gap with the slower teacher RNNG, which benefits from syntactic bias and annotation. 
These findings confirm our hypothesis that the KD approach constitutes an efficient way to inject hierarchical bias into sequential models.
\vspace{-1mm}
\item The born-again model (BA-LSTM) outperforms the LSTM LM, albeit by a small margin. This finding suggests that KD helps improve the syntactic competence of LSTMs, even when the teacher model lacks explicit hierarchical bias and shares the same architecture as the student.
\vspace{-1mm}
\item In terms of perplexity, both BA-LSTM and DSA-LSTM perform slightly worse than the full LSTM LM trained without KD loss. We attribute this gap to the smoother target distribution when using KD, which effectively penalises high probabilities on the correct next word $x^{*}_j$ unless the teacher model is extremely confident. This observation is consistent with earlier findings on label smoothing in machine translation \cite{pereyra_2017,vaswani_2017}, which often results in better BLEU at the expense of slightly worse likelihood.
\vspace{-1mm}
\item Despite identical architectures, on aggregate the DSA-LSTM substantially improves over the full LSTM (85\% to 89\%), constituting a 27\% error rate reduction and a new state of the art. Our findings suggest that the DSA-LSTM combines the best of both hierarchical bias and data scale: on subject-verb agreement, the DSA-LSTM improves over the LSTM baseline and narrows the gap with the teacher RNNG, while at the same time performing well on reflexive anaphora and NPI, on which the teacher RNNG (but not the full LSTM) fails to achieve a good performance.
\vspace{-1mm}
\item While not directly comparable, the DSA-LSTM outperforms a pre-trained BERT \cite{devlin_2019,goldberg_2019}\footnote{\newcite{goldberg_2019} applies an additional pre-processing step, removing sentences in which the focus verb does not appear as a single word in the word piece-based vocabulary; hence, the evaluation sentences are slightly different.} on subject-verb agreement. Since BERT benefits from bidirectionality and was trained on 30 times as much data as the DSA-LSTM, this finding suggests that, at least in terms of syntactic competence, structural biases continue to be relevant even as the current generation of sequential LMs is able to exploit increasingly large amounts of data.
\end{itemizesquish}
\vspace{-2mm}
\subsection{Probing for Hierarchical Information}
Having established the advantages of the DSA-LSTM on targeted syntactic evaluations, we turn to the question of analysing how its internal representation differs from that of a standard LSTM LM. To this end, we adopt the method of \newcite{blevins_2018} and use a probe \cite[\emph{inter alia}]{shi_16,adi_2017,belinkov_2017a,conneau_2018,hewitt_2019} that predicts the \emph{grandparent constituent} of a word token $x_t$, based on its encoding $\mathbf{h}_t$ under the pre-trained LSTM. Under this framework, the accuracy of the probe on a held-out set can be understood as an indication of how well the hidden states encode the relevant syntactic information required to succeed in this task. 

We use a linear classifier for the probe and obtain the predicted grandparent constituent label using the same pre-trained Berkeley parser (\S3) that we used to obtain predicted phrase-structure trees to train the RNNG. For the probing experiment, we randomly select sentences from each respective training, validation, and test set of the \newcite{gulordava_18} dataset to yield $\sim$300,000 words for training and $\sim$10,000 words for each of validation and test sets. For the probe features, we use a concatenation of the LSTM hidden state at the current and next words,\footnote{Our probing feature set thus slightly differs from that of \newcite{blevins_2018}, who concatenated the hidden states of a left-to-right and right-to-left LSTM language models.} i.e. $[\mathbf{h}_t; \mathbf{h}_{t+1}]$, where $;$ denotes the concatenation operation.   

\begin{figure}[t]
    \centering
    \includegraphics[scale=0.38]{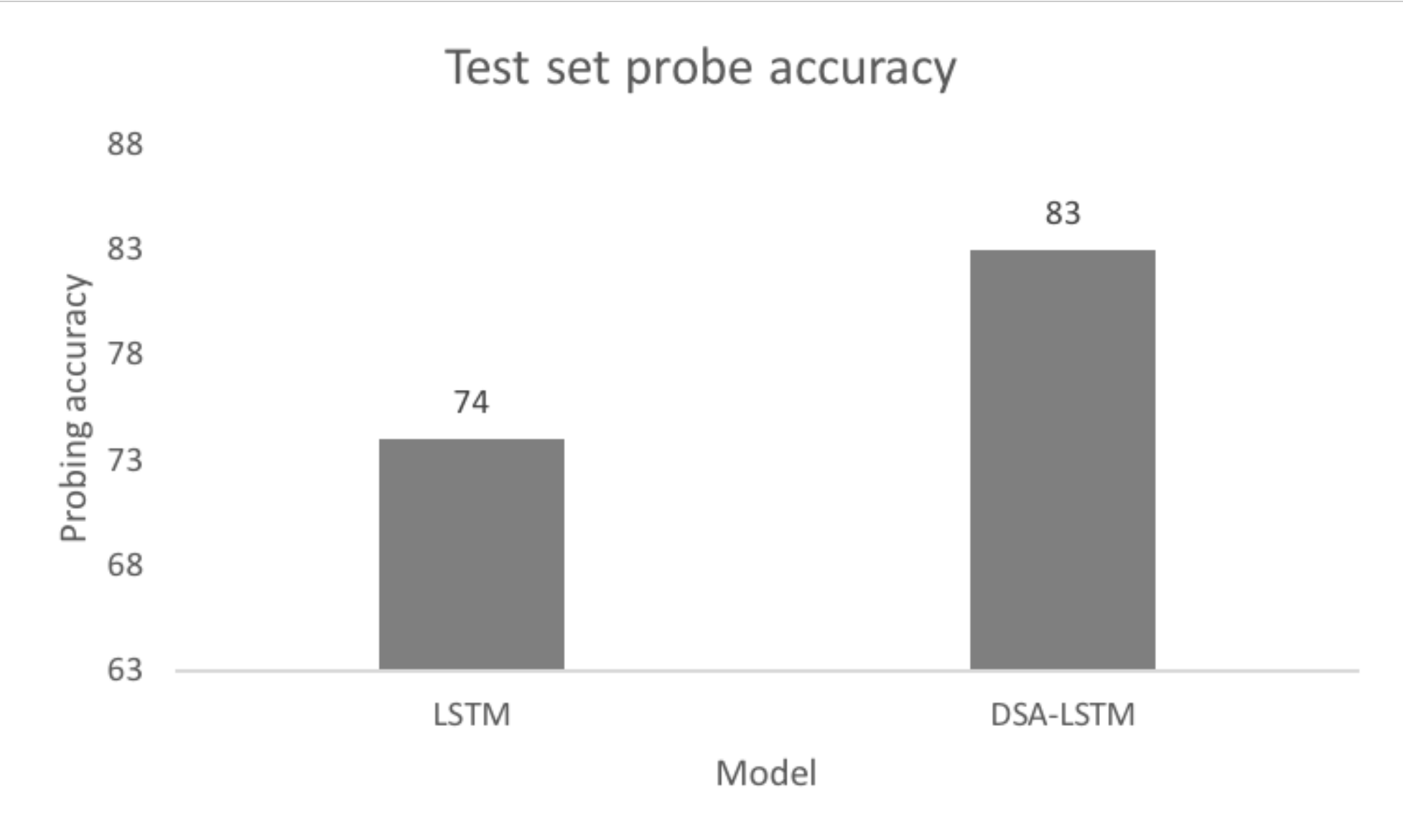}
    \vspace{-2mm}
    \caption{Probing accuracy on the test set. We analyse the hidden states of the LSTM and DSA-LSTM to analyse the structural information encoded in each respective model's hidden state.}
    \label{fig:probe}
     \vspace{-4mm}
\end{figure}

Recall that the DSA-LSTM operates only on word sequences and has no access to the Berkeley parse during training. We summarise the probing result in Fig.~\ref{fig:probe}. Overall, the syntactic probing accuracy for the DSA-LSTM is much higher than for the LSTM LM (83\% to 74\%; a 34\% error rate reduction), suggesting that the means by which the DSA-LSTM achieves better syntactic competence is by tracking more hierarchical information during sequential processing.

\section{Related Work}
Augmenting language models with syntactic information and structural inductive bias has been a long-standing area of research. To this end, syntactic language models estimate the joint probability of surface strings and some form of syntactic structure \cite{jurafsky_1995,chelba:2000,roark:2001,henderson:2004,emami:2005,buys:2015a,mirowski_2015,rnng,kim_2019}. In contrast to these approaches, the DSA-LSTM only models the probability of surface strings, albeit with an auxiliary loss that distills the next-word predictive distribution of a syntactic language model. 

Earlier work has also explored multi-task learning with syntactic objectives as an auxiliary loss in language modelling and machine translation \cite{luong_2016,eriguchi_2016,nadejde_2017,enguehard_2017,aharoni_2017,eriguichi_2017}. Our approach of injecting syntactic bias through a KD objective is orthogonal to this approach, with the primary difference that here the student DSA-LSTM has no direct access to syntactic annotations; it does, however, have access to the teacher RNNG's softmax distribution over the next word. 

Our approach is also closely related to recent work that introduces structurally-motivated inductive biases into language models. \newcite{chung_2016} segmented the hidden state update of an RNN through a multi-scale hierarchical recurrence, thereby providing a shortcut to the gradient propagation of long-range, hierarchical dependencies. \newcite{yogatama:2018} introduced a stack-structured memory to encourage hierarchical modelling in language models, where the resulting model successfully outperforms standard LSTM variants in number agreement \cite{linzen-2016} evaluation. \newcite{shen_2019} imposed a hierarchical bias on the LSTM cell-updating mechanism, based on the intuition that larger constituents contain information that changes more slowly across the sequence. Our proposed method is orthogonal and can be applied on top of these recent approaches.   
\ignore{
\begin{abstract}
As language exhibits hierarchical structure, we address the problem of designing language models that benefit from hierarchical bias and can capture complex syntactic dependencies to a large extent, albeit without introducing much overhead in terms of scalability. To this end, we employ knowledge distillation (KD) as a means of injecting hierarchical bias from a slow syntactic language model that leverages syntactic information, to a more scalable student LSTM that operates sequentially. While we find that LSTM language models are able to achieve better syntactic generalisation than previously thought, our proposed combination of hierarchical bias and scalability improves over this strong baseline and yields a new state of the art on targeted syntactic evaluations. Analysis suggests that, compared to LSTM language models, our approach has better sample complexity for capturing complex syntactic constructions and encodes hierarchical information to a large extent, despite lacking direct access to syntactic annotation.  
\end{abstract}
}
\section{Conclusion}
\vspace{-2mm}
In this paper, we introduce a distilled syntax-aware LSTM (DSA-LSTM), which combines scalability with structural biases. We achieve this by distilling the predictions about upcoming words in a large training corpus made by a (computationally complex) hierarchical language model trained on a small subset of the data. While we find that LSTM language models achieve better syntactic generalisation than previously thought, on targeted syntactic evaluations our approach improves over this strong baseline, yields a new state of the art, compares favourably to a language model trained on much more data, and results in a language model that encodes hierarchical information to a large extent despite its sequential architecture. Our approach is a general one that can be applied to other student model architectures, such as Transformers \cite{vaswani_2017}. These findings suggest that the question of structural biases continues to be relevant for improving syntactic competence, even in scalable architectures that can benefit from ever-growing amounts of training data.

\section*{Acknowledgments}
We would like to thank Rebecca Marvin and Tal Linzen for their help in answering questions regarding data preparation. We also thank Dani Yogatama, John Hale, and the three anonymous reviewers for their helpful suggestions.\\


\bibliography{acl2019}

\begin{thebibliography}{60}
\expandafter\ifx\csname natexlab\endcsname\relax\def\natexlab#1{#1}\fi

\bibitem[{Adi et~al.(2017)Adi, Kermany, Belinkov, Lavi, and
  Goldberg}]{adi_2017}
Yossi Adi, Einat Kermany, Yonatan Belinkov, Ofer Lavi, and Yoav Goldberg. 2017.
\newblock Fine-grained analysis of sentence embeddings using auxiliary
  prediction tasks.
\newblock In \emph{Proc. of ICLR}.

\bibitem[{Aharoni and Goldberg(2017)}]{aharoni_2017}
Roee Aharoni and Yoav Goldberg. 2017.
\newblock Towards string-to-tree neural machine translation.
\newblock In \emph{Proc. of ACL}.

\bibitem[{Ba and Caruana(2014)}]{ba_et_al_14}
Jimmy Ba and Rich Caruana. 2014.
\newblock Do deep nets really need to be deep?
\newblock In \emph{NIPS}.

\bibitem[{Bangalore and Joshi(1999)}]{bangalore_1999}
Srinivas Bangalore and Aravind~K. Joshi. 1999.
\newblock Supertagging: An approach to almost parsing.
\newblock \emph{Computational Linguistics}, 25.

\bibitem[{Belinkov et~al.(2017)Belinkov, Durrani, Dalvi, Sajjad, and
  Glass}]{belinkov_2017a}
Yonatan Belinkov, Nadir Durrani, Fahim Dalvi, Hassan Sajjad, and James Glass.
  2017.
\newblock What do neural machine translation models learn about morphology?
\newblock In \emph{Proc. of ACL}.

\bibitem[{Blevins et~al.(2018)Blevins, Levy, and and}]{blevins_2018}
Terra Blevins, Omer Levy, and Luke~Zettlemoyer and. 2018.
\newblock Deep rnns encode soft hierarchical syntax.
\newblock In \emph{Proc. of ACL}.

\bibitem[{Bucil\v{a} et~al.(2006)Bucil\v{a}, Caruana, and
  Niculescu-Mizil}]{bucila:2006}
Cristian Bucil\v{a}, Rich Caruana, and Alexandru Niculescu-Mizil. 2006.
\newblock Model compression.
\newblock In \emph{Proc. of KDD}.

\bibitem[{Buys and Blunsom(2015)}]{buys:2015a}
Jan Buys and Phil Blunsom. 2015.
\newblock A {Bayesian} model for generative transition-based dependency
  parsing.
\newblock \emph{CoRR}, abs/1506.04334.

\bibitem[{Chelba and Jelinek(2000)}]{chelba:2000}
Ciprian Chelba and Frederick Jelinek. 2000.
\newblock Structured language modeling.
\newblock \emph{Computer Speech and Language}, 14(4).

\bibitem[{Choe and Charniak(2016)}]{choe:2016}
Do~Kook Choe and Eugene Charniak. 2016.
\newblock Parsing as language modeling.
\newblock In \emph{Proc. of EMNLP}.

\bibitem[{Chung et~al.(2017)Chung, Ahn, and Bengio}]{chung_2016}
Junyoung Chung, Sungjin Ahn, and Yoshua Bengio. 2017.
\newblock Hierarchical multiscale recurrent neural networks.
\newblock In \emph{Proc. of ICLR}.

\bibitem[{Clark and Curran(2007)}]{clark-curran-2007-wide}
Stephen Clark and James~R. Curran. 2007.
\newblock Wide-coverage efficient statistical parsing with {CCG} and log-linear
  models.
\newblock \emph{Computational Linguistics}.

\bibitem[{Conneau et~al.(2018)Conneau, Kruszewski, Lample, Barrault, and
  Baroni}]{conneau_2018}
Alexis Conneau, Germ{\'{a}}n Kruszewski, Guillaume Lample, Lo{\"{\i}}c
  Barrault, and Marco Baroni. 2018.
\newblock What you can cram into a single vector: Probing sentence embeddings
  for linguistic properties.

\bibitem[{Dehghani et~al.(2019)Dehghani, Gouws, Vinyals, Uszkoreit, and
  Kaiser}]{dehghani_19}
Mostafa Dehghani, Stephan Gouws, Oriol Vinyals, Jakob Uszkoreit, and Lukasz
  Kaiser. 2019.
\newblock Universal transformers.
\newblock In \emph{Proc. of ICLR}.

\bibitem[{Devlin et~al.(2019)Devlin, Chang, Lee, and Toutanova}]{devlin_2019}
Jacob Devlin, Ming{-}Wei Chang, Kenton Lee, and Kristina Toutanova. 2019.
\newblock {BERT:} pre-training of deep bidirectional transformers for language
  understanding.
\newblock In \emph{Proc. of NAACL}.

\bibitem[{Dyer et~al.(2015)Dyer, Ballesteros, Ling, Matthews, and
  Smith}]{stack_lstm}
Chris Dyer, Miguel Ballesteros, Wang Ling, Austin Matthews, and Noah~A. Smith.
  2015.
\newblock Transition-based dependency parsing with stack long short-term
  memory.
\newblock In \emph{Proc. of ACL}.

\bibitem[{Dyer et~al.(2016)Dyer, Kuncoro, Ballesteros, and Smith}]{rnng}
Chris Dyer, Adhiguna Kuncoro, Miguel Ballesteros, and Noah~A. Smith. 2016.
\newblock Recurrent neural network grammars.
\newblock In \emph{Proc. of NAACL}.

\bibitem[{Emami and Jelinek(2005)}]{emami:2005}
Ahmad Emami and Frederick Jelinek. 2005.
\newblock A neural syntactic language model.
\newblock \emph{Machine Learning}, 60:195--227.

\bibitem[{Enguehard et~al.(2017)Enguehard, Goldberg, and
  Linzen}]{enguehard_2017}
{\'E}mile Enguehard, Yoav Goldberg, and Tal Linzen. 2017.
\newblock Exploring the syntactic abilities of rnns with multi-task learning.
\newblock In \emph{Proc. of CoNLL}.

\bibitem[{Eriguchi et~al.(2016)Eriguchi, Hashimoto, and
  Tsuruoka}]{eriguchi_2016}
Akiko Eriguchi, Kazuma Hashimoto, and Yoshimasa Tsuruoka. 2016.
\newblock Tree-to-sequence attentional neural machine translation.
\newblock In \emph{Proc. of ACL}.

\bibitem[{Eriguchi et~al.(2017)Eriguchi, Tsuruoka, and Cho}]{eriguichi_2017}
Akiko Eriguchi, Yoshimasa Tsuruoka, and Kyunghyun Cho. 2017.
\newblock Learning to parse and translate improves neural machine translation.
\newblock In \emph{Proc. of ACL}.

\bibitem[{Fried et~al.(2017)Fried, Stern, and Klein}]{fried_17}
Daniel Fried, Mitchell Stern, and Dan Klein. 2017.
\newblock Improving neural parsing by disentangling model combination and
  reranking effects.
\newblock In \emph{Proc. of ACL}.

\bibitem[{Furlanello et~al.(2018)Furlanello, Lipton, Tschannen, Itti, and
  Anandkumar}]{furlanello_18}
Tommaso Furlanello, Zachary~Chase Lipton, Michael Tschannen, Laurent Itti, and
  Anima Anandkumar. 2018.
\newblock Born-again neural networks.
\newblock In \emph{Proc. of ICML}.

\bibitem[{Goldberg(2019)}]{goldberg_2019}
Yoav Goldberg. 2019.
\newblock Assessing {BERT}'s syntactic abilities.
\newblock \emph{CoRR}, abs/1901.05287.

\bibitem[{Gulordava et~al.(2018)Gulordava, Bojanowski, Grave, Linzen, and
  Baroni}]{gulordava_18}
Kristina Gulordava, Piotr Bojanowski, Edouard Grave, Tal Linzen, and Marco
  Baroni. 2018.
\newblock Colorless green recurrent networks dream hierarchically.
\newblock In \emph{Proc. of NAACL}.

\bibitem[{Hale et~al.(2018)Hale, Dyer, Kuncoro, and Brennan}]{hale_2018}
John Hale, Chris Dyer, Adhiguna Kuncoro, and Jonathan~R. Brennan. 2018.
\newblock Finding syntax in human encephalography with beam search.
\newblock In \emph{Proc. of ACL}.

\bibitem[{Henderson(2004)}]{henderson:2004}
James Henderson. 2004.
\newblock Discriminative training of a neural network statistical parser.
\newblock In \emph{Proc. of ACL}.

\bibitem[{Hewitt and Manning(2019)}]{hewitt_2019}
John Hewitt and Christopher~D. Manning. 2019.
\newblock {A} structural probe for finding syntax in word representations.
\newblock In \emph{Proc. of NAACL}.

\bibitem[{Hinton et~al.(2015)Hinton, Vinyals, and Dean}]{dark_knowledge}
Geoffrey~E. Hinton, Oriol Vinyals, and Jeffrey Dean. 2015.
\newblock Distilling the knowledge in a neural network.
\newblock \emph{CoRR}, abs/1503.02531.

\bibitem[{Hochreiter and Schmidhuber(1997)}]{hochreiter_97}
Sepp Hochreiter and J\"{u}rgen Schmidhuber. 1997.
\newblock Long short-term memory.
\newblock \emph{Neural Computation}.

\bibitem[{Howard and Ruder(2018)}]{howard_2018}
Jeremy Howard and Sebastian Ruder. 2018.
\newblock Universal language model fine-tuning for text classification.
\newblock In \emph{Proc. of ACL}.

\bibitem[{Jelinek(1997)}]{jelinek_1998}
Frederick Jelinek. 1997.
\newblock \emph{Statistical Methods for Speech Recognition}.
\newblock MIT Press.

\bibitem[{{Jurafsky} et~al.(1995){Jurafsky}, {Wooters}, {Segal}, {Stolcke},
  {Fosler}, {Tajchaman}, and {Morgan}}]{jurafsky_1995}
D.~{Jurafsky}, C.~{Wooters}, J.~{Segal}, A.~{Stolcke}, E.~{Fosler},
  G.~{Tajchaman}, and N.~{Morgan}. 1995.
\newblock \href {https://doi.org/10.1109/ICASSP.1995.479396} {Using a
  stochastic context-free grammar as a language model for speech recognition}.
\newblock In \emph{Proc. of ICASSP}.

\bibitem[{Kim and Rush(2016)}]{seq-distillation}
Yoon Kim and Alexander~M. Rush. 2016.
\newblock Sequence-level knowledge distillation.
\newblock In \emph{Proc. of EMNLP}.

\bibitem[{Kim et~al.(2019)Kim, Rush, Yu, Kuncoro, Dyer, and Melis}]{kim_2019}
Yoon Kim, Alexander~M. Rush, Lei Yu, Adhiguna Kuncoro, Chris Dyer, and Gabor
  Melis. 2019.
\newblock Unsupervised recurrent neural network grammars.
\newblock In \emph{Proc. of NAACL}.

\bibitem[{Koehn(2010)}]{koehn_2010}
Philipp Koehn. 2010.
\newblock \emph{Statistical Machine Translation}.
\newblock Cambridge University Press.

\bibitem[{Kuncoro et~al.(2017)Kuncoro, Ballesteros, Kong, Dyer, Neubig, and
  Smith}]{kuncoro-2017}
Adhiguna Kuncoro, Miguel Ballesteros, Lingpeng Kong, Chris Dyer, Graham Neubig,
  and Noah~A. Smith. 2017.
\newblock What do recurrent neural network grammars learn about syntax?
\newblock In \emph{Proc. of EACL}.

\bibitem[{Kuncoro et~al.(2018)Kuncoro, Dyer, Hale, Yogatama, Clark, and
  Blunsom}]{kuncoro_2018}
Adhiguna Kuncoro, Chris Dyer, John Hale, Dani Yogatama, Stephen Clark, and Phil
  Blunsom. 2018.
\newblock Lstms can learn syntax-sensitive dependencies well, but modeling
  structure makes them better.
\newblock In \emph{Proc. of ACL}.

\bibitem[{Linzen et~al.(2016)Linzen, Dupoux, and Goldberg}]{linzen-2016}
Tal Linzen, Emmanuel Dupoux, and Yoav Goldberg. 2016.
\newblock Assessing the ability of {LSTMs} to learn syntax-sensitive
  dependencies.
\newblock \emph{Transactions of the Association for Computational Linguistics}.

\bibitem[{Liu et~al.(2018)Liu, Che, Zhao, Qin, and Liu}]{liu_2018}
Yijia Liu, Wanxiang Che, Huaipeng Zhao, Bing Qin, and Ting Liu. 2018.
\newblock Distilling knowledge for search-based structured prediction.
\newblock In \emph{Proc. of ACL}.

\bibitem[{Luong et~al.(2016)Luong, Le, Sutskever, Vinyals, and
  Kaiser}]{luong_2016}
Thang Luong, Quoc~V. Le, Ilya Sutskever, Oriol Vinyals, and Lukasz Kaiser.
  2016.
\newblock Multi-task sequence to sequence learning.
\newblock In \emph{Proc. of ICLR}.

\bibitem[{Marvin and Linzen(2018)}]{marvin_2018}
Rebecca Marvin and Tal Linzen. 2018.
\newblock Targeted syntactic evaluation of language models.
\newblock In \emph{Proc. of EMNLP}.

\bibitem[{Mikolov et~al.(2010)Mikolov, Karafi\'at, Burget, \v{C}ernock\'y, and
  Khudanpur}]{mikolov:2010}
Tom\'a\v{s} Mikolov, Martin Karafi\'at, Luk\'a\v{s} Burget, Jan \v{C}ernock\'y,
  and Sanjeev Khudanpur. 2010.
\newblock Recurrent neural network based language model.
\newblock In \emph{Proc. of Interspeech}.

\bibitem[{Mirowski and Vlachos(2015)}]{mirowski_2015}
Piotr Mirowski and Andreas Vlachos. 2015.
\newblock Dependency recurrent neural language models for sentence completion.
\newblock In \emph{Proc. of ACL-IJCNLP}.

\bibitem[{Nadejde et~al.(2017)Nadejde, Reddy, Sennrich, Dwojak,
  Junczys-Dowmunt, Koehn, and Birch}]{nadejde_2017}
Maria Nadejde, Siva Reddy, Rico Sennrich, Tomasz Dwojak, Marcin
  Junczys-Dowmunt, Philipp Koehn, and Alexandra Birch. 2017.
\newblock Predicting target language ccg supertags improves neural machine
  translation.
\newblock In \emph{Proc. of WMT}.

\bibitem[{Neubig et~al.(2017{\natexlab{a}})Neubig, Dyer, Goldberg, Matthews,
  Ammar, Anastasopoulos, Ballesteros, Chiang, Clothiaux, Cohn, Duh, Faruqui,
  Gan, Garrette, Ji, Kong, Kuncoro, Kumar, Malaviya, Michel, Oda, Richardson,
  Saphra, Swayamdipta, and Yin}]{dynet}
Graham Neubig, Chris Dyer, Yoav Goldberg, Austin Matthews, Waleed Ammar,
  Antonios Anastasopoulos, Miguel Ballesteros, David Chiang, Daniel Clothiaux,
  Trevor Cohn, Kevin Duh, Manaal Faruqui, Cynthia Gan, Dan Garrette, Yangfeng
  Ji, Lingpeng Kong, Adhiguna Kuncoro, Gaurav Kumar, Chaitanya Malaviya, Paul
  Michel, Yusuke Oda, Matthew Richardson, Naomi Saphra, Swabha Swayamdipta, and
  Pengcheng Yin. 2017{\natexlab{a}}.
\newblock {DyNet: The Dynamic Neural Network Toolkit}.
\newblock \emph{arXiv preprint arXiv:1701.03980}.

\bibitem[{Neubig et~al.(2017{\natexlab{b}})Neubig, Goldberg, and
  Dyer}]{neubig_2017}
Graham Neubig, Yoav Goldberg, and Chris Dyer. 2017{\natexlab{b}}.
\newblock On-the-fly operation batching in dynamic computation graphs.
\newblock In \emph{Proc. of NIPS}.

\bibitem[{Pereyra et~al.(2017)Pereyra, Tucker, Chorowski, Kaiser, and
  Hinton}]{pereyra_2017}
Gabriel Pereyra, George Tucker, Jan Chorowski, Lukasz Kaiser, and Geoffrey
  Hinton. 2017.
\newblock Regularizing neural networks by penalizing confident output
  distributions.
\newblock In \emph{Proc. of ICLR}.

\bibitem[{Peters et~al.(2018)Peters, Neumann, Iyyer, Gardner, Clark, Lee, and
  Zettlemoyer}]{peters_2018}
Matthew~E. Peters, Mark Neumann, Mohit Iyyer, Matt Gardner, Christopher Clark,
  Kenton Lee, and Luke Zettlemoyer. 2018.
\newblock Deep contextualized word representations.
\newblock In \emph{Proc. of NAACL}.

\bibitem[{Petrov and Klein(2007)}]{petrov:2007}
Slav Petrov and Dan Klein. 2007.
\newblock Improved inference for unlexicalized parsing.
\newblock In \emph{Proc. of NAACL}.

\bibitem[{Radford et~al.(2019)Radford, Wu, Child, Luan, Amodei, and
  Sutskever}]{radford_2019}
Alec Radford, Jeffrey Wu, Rewon Child, David Luan, Dario Amodei, and Ilya
  Sutskever. 2019.
\newblock Language models are unsupervised multitask learners.

\bibitem[{Roark(2001)}]{roark:2001}
Brian Roark. 2001.
\newblock Probabilistic top-down parsing and language modeling.
\newblock \emph{Computational Linguistics}, 27(2).

\bibitem[{Rosenfeld(2000)}]{rosenfeld_2000}
Ronald Rosenfeld. 2000.
\newblock Two decades of statistical language modeling: Where do we go from
  here.
\newblock In \emph{Proc. of IEEE}.

\bibitem[{Shen et~al.(2019)Shen, Tan, Sordoni, and Courville}]{shen_2019}
Yikang Shen, Shawn Tan, Alessandro Sordoni, and Aaron~C. Courville. 2019.
\newblock Ordered neurons: Integrating tree structures into recurrent neural
  networks.
\newblock In \emph{Proc. of ICLR}.

\bibitem[{Shi et~al.(2016)Shi, Padhi, and Knight}]{shi_16}
Xing Shi, Inkit Padhi, and Kevin Knight. 2016.
\newblock Does string-based neural {MT} learn source syntax?
\newblock In \emph{Proc. of EMNLP}.

\bibitem[{Stern et~al.(2017)Stern, Fried, and Klein}]{stern_17}
Mitchell Stern, Daniel Fried, and Dan Klein. 2017.
\newblock Effective inference for generative neural parsing.
\newblock In \emph{Proc. of EMNLP}.

\bibitem[{Szegedy et~al.(2016)Szegedy, Vanhoucke, Ioffe, Shlens, and
  Wojna}]{szegedy_2016}
Christian Szegedy, Vincent Vanhoucke, Sergey Ioffe, Jonathon Shlens, and
  Zbigniew Wojna. 2016.
\newblock Rethinking the inception architecture for computer vision.
\newblock In \emph{Proc. of CVPR}.

\bibitem[{Tran et~al.(2018)Tran, Bisazza, and Monz}]{tran_2018}
Ke~M. Tran, Arianna Bisazza, and Christof Monz. 2018.
\newblock The importance of being recurrent for modeling hierarchical
  structure.
\newblock In \emph{Proc. of EMNLP}.

\bibitem[{Vaswani et~al.(2017)Vaswani, Shazeer, Parmar, Uszkoreit, Jones,
  Gomez, Kaiser, and Polosukhin}]{vaswani_2017}
Ashish Vaswani, Noam Shazeer, Niki Parmar, Jakob Uszkoreit, Llion Jones,
  Aidan~N. Gomez, Lukasz Kaiser, and Illia Polosukhin. 2017.
\newblock Attention is all you need.
\newblock In \emph{Proc. of NIPS}.

\bibitem[{Yogatama et~al.(2018)Yogatama, Miao, Melis, Ling, Kuncoro, Dyer, and
  Blunsom}]{yogatama:2018}
Dani Yogatama, Yishu Miao, Gabor Melis, Wang Ling, Adhiguna Kuncoro, Chris
  Dyer, and Phil Blunsom. 2018.
\newblock Memory architectures in recurrent neural network language models.
\newblock In \emph{Proc. of ICLR}.

\end{thebibliography}
\bibliographystyle{acl_natbib}

\clearpage

\appendix

\section*{Appendix}
Here we outline the hyperparameters and the experimental results with standard deviation values. 
\vspace{-2mm}
\section{Hyperparameters}
The hyperparameters for each model is summarised as follows.
\vspace{-2mm}
\paragraph{LSTM LMs.} For the LSTM LMs trained on the full and small training sets, we use the following hyperparameters that achieve the best validation perplexity following a grid search: 2-layer LSTM with 650 hidden units per layer for the full LSTM and 300 hidden units per layer for the small LSTM (similar model capacity as the RNNG trained on the same smaller training set), optimised by stochastic gradient descent (SGD) with a learning rate of 0.45 (decayed exponentially at every epoch with a factor of 0.9 after the tenth epoch), a dropout rate of 0.2 applied on both input and recurrent connections, and a batch size of 20 sentences. 
\vspace{-2mm}
\paragraph{RNNG.} For the RNNG, we use the following hyperparameters that achieve the best validation perplexity following a similar grid search: 2-layer stack LSTM with 256 hidden units per layer, optimised by SGD with a learning rate of 0.3 (decayed exponentially at every epoch with a factor of 0.92 after the tenth epoch), a dropout rate of 0.3 applied on both input and recurrent connections, and a batch size of 10 sentences.
\vspace{-2mm}
\paragraph{DSA-LSTMs and Born-Again LSTMs.} We use the same hyperparameters for the DSA-LSTMs trained on both the full and small (S-DSA-LSTM) training sets and the born-again LSTM (BA-LSTM) trained on the full training set. Since the model architectures are identical with the respective LSTM LMs (i.e. only the training objective is different), we only optimise for the learning rates and keep all other hyperparameters the same. We find that a learning rate of 0.4 and an exponential decay factor of 0.9 applied after the tenth epoch works well across all three models trained with the KD objective.

\section{Experimental Results with Standard Deviation}
We summarise the experimental results that include standard deviation values in Table~\ref{tab:full_table}. 

\begin{table*}[!htb]
      \centering
      \resizebox{\textwidth}{!}{%
       \begin{tabular}{l|r|r||r|r|r||r||r}
       & \multicolumn{2}{c||}{\textbf{Small Training Set}} & \multicolumn{3}{c||}{\textbf{Full Training Set}} \\ \hline
		& \textbf{S-DSA-LSTM}$^{\dagger}$ &  \textbf{RNNG}$^{\dagger}$ & \textbf{Full LSTM} & \textbf{BA-LSTM} & \textbf{DSA-LSTM} &\textbf{BERT} & \textbf{Humans}  \\
		\hline\hline
		Gulordava et al. (2018) \textbf{test ppl.} & 93.95 ($\pm$0.18)  & \textbf{92.30} ($\pm$0.27) & \textbf{53.73} ($\pm$0.16)  &  54.64 ($\pm$0.25) &  56.74 ($\pm$0.26) & N/A & N/A \\ \hline \hline
	    \multicolumn{8}{c}{SUBJECT-VERB AGREEMENT}  \\ \hline
		Simple & 0.96 ($\pm$0.03)  & \textbf{0.99} ($\pm$0.01) & \textbf{1.00} ($\pm$0.00)  & \textbf{1.00} ($\pm$0.00)  & \textbf{1.00} ($\pm$0.00) & 1.00 & 0.96 \\ 
        In a sentential complement & \textbf{0.98} ($\pm$0.02)  & 0.93 ($\pm$0.02) & 0.97 ($\pm$0.02) & \textbf{0.98} ($\pm$0.02) & \textbf{0.98} ($\pm$0.02) & 0.83 & 0.93 \\ 
        Short VP coordination &  0.88 ($\pm$0.04) & \textbf{0.96} ($\pm$0.02) & 0.96 ($\pm$0.02) & 0.95 ($\pm$0.02)  & \textbf{0.99} ($\pm$0.02)   & 0.89 & 0.94 \\ 
        Long VP coordination & 0.74 ($\pm$0.03) & \textbf{0.94} ($\pm$0.03) & \textbf{0.82} ($\pm$0.05) & 0.80 ($\pm$0.04)  & 0.80 ($\pm$0.02)   & 0.98 & 0.82 \\ 
        Across a prepositional phrase & 0.88 ($\pm$0.02)   & \textbf{0.95} ($\pm$0.01) & 0.89 ($\pm$0.02)  & 0.89 ($\pm$0.03)  & \textbf{0.91} ($\pm$0.03)  & 0.85 & 0.85 \\ 
        Across a subject relative clause & 0.87 ($\pm$0.02) & \textbf{0.95} ($\pm$0.03) & 0.87 ($\pm$0.02)  & 0.87 ($\pm$0.01)  & \textbf{0.90} ($\pm$0.02) & 0.84 & 0.88 \\
        Across an object relative clause & 0.69 ($\pm$0.06) & \textbf{0.95} ($\pm$0.03) & 0.77 ($\pm$0.11)  &  0.81 ($\pm$0.05) & \textbf{0.84} ($\pm$0.03) & 0.89 & 0.85 \\
        Across an object relative clause (no \emph{that}) & 0.61 ($\pm$0.05) & \textbf{0.93} ($\pm$0.02)  & 0.70 ($\pm$0.05)  & 0.74 ($\pm$0.03)  & \textbf{0.77} ($\pm$0.02) & 0.86 & 0.82 \\
        In an object relative clause & 0.87 ($\pm$0.05) & \textbf{0.96} ($\pm$0.01) & 0.90 ($\pm$0.03)  & 0.91 ($\pm$0.03)  & \textbf{0.92} ($\pm$0.04) & 0.95 & 0.78 \\
        In an object relative clause (no \emph{that}) & 0.88 ($\pm$0.03) & \textbf{0.96} ($\pm$0.02) & 0.86 ($\pm$0.05)  & 0.83 ($\pm$0.02)  & \textbf{0.92} ($\pm$0.02) & 0.79 & 0.79 \\ \hline
         \textbf{Average of subject-verb agreement} & 0.84 ($\pm$0.02)  & \textbf{0.95} ($\pm$0.01) & 0.87 ($\pm$0.02) & 0.88 ($\pm$0.01)  & \textbf{0.90} ($\pm$0.01) & 0.89 & 0.86 \\ \hline \hline
        \multicolumn{8}{c}{REFLEXIVE ANAPHORA}  \\ \hline
        Simple & \textbf{0.90} ($\pm$0.01) & 0.83 ($\pm$0.02) & 0.91 ($\pm$0.01)  & \textbf{0.92} ($\pm$0.03)  & 0.91 ($\pm$0.04) & 0.94 & 0.96\\ 
       In a sentential complement & \textbf{0.78} ($\pm$0.01)  & 0.46 ($\pm$0.05) & 0.81 ($\pm$0.02)  & 0.81 ($\pm$0.02)  & \textbf{0.82} ($\pm$0.03) & 0.89 & 0.91  \\ 
       Across a relative clause & 0.67 ($\pm$0.03)  & \textbf{0.82} ($\pm$0.02) & 0.64 ($\pm$0.02)  & 0.64 ($\pm$0.02)  & \textbf{0.67} ($\pm$0.03) & 0.80 & 0.87  \\ \hline
        \textbf{Average of reflexive anaphora} & \textbf{0.78} ($\pm$0.01)  & 0.70 ($\pm$0.02) & 0.79 ($\pm$0.01)   & 0.79 ($\pm$0.02)  & \textbf{0.80} ($\pm$0.03) & 0.88 & 0.91 \\ \hline \hline
       \multicolumn{8}{c}{NEGATIVE POLARITY ITEMS}  \\ \hline
       Simple & \textbf{0.84} ($\pm$0.05) & 0.28 ($\pm$0.05) & 0.96 ($\pm$0.04)  & \textbf{0.98} ($\pm$0.02)  & 0.94 ($\pm$0.04) & N/A & 0.98 \\ 
       Across a relative clause & 0.73 ($\pm$0.07)  & \textbf{0.78} ($\pm$0.06) & 0.75 ($\pm$0.12) & 0.70 ($\pm$0.10)  & \textbf{0.91} ($\pm$0.07) & N/A &  0.81  \\ \hline
        \textbf{Average of negative polarity items} & \textbf{0.79} ($\pm$0.05)  & 0.53 ($\pm$0.04) & 0.86 ($\pm$0.06)  & 0.84 ($\pm$0.05)  & \textbf{0.92} ($\pm$0.05) & N/A & 0.90 \\ \hline \hline
       \textbf{Average of all constructions} & 0.82 ($\pm$0.02)  & \textbf{0.85} ($\pm$0.02) & 0.85 ($\pm$0.02)  & 0.86 ($\pm$0.01)  & \textbf{0.89} ($\pm$0.01) & N/A &  0.88  
		\end{tabular}}
		\caption{Experimental findings of the \textbf{``DSA-LSTM''}. For each column, we report the mean and standard deviation values of 10 identical models trained from different random seeds. \textbf{``S-DSA-LSTM''} indicates the DSA-LSTM trained on the smaller RNNG training set, while \textbf{``BA-LSTM''} is the born-again model where the teacher is the full LSTM LM. We also compare with the syntactic generalisation of \textbf{``BERT''} Base, which is not strictly comparable since it is trained on 30 times as much data. $^{\dagger}$ indicates models trained on the smaller 20\% training set (\S\ref{sec:rnng}). Results in bold denote the best among those trained with the same amounts of data.}
        \label{tab:full_table}
\end{table*}

\end{document}